%% file: eacl2023.tex
\pdfoutput=1

\documentclass[11pt]{article}

\usepackage{EACL2023}

\usepackage{times}
\usepackage{latexsym}

\usepackage[T1]{fontenc}

\usepackage[utf8]{inputenc}

\usepackage{microtype}

\usepackage{inconsolata}

\usepackage{graphicx}
\usepackage{amsmath}

\title{Automated Query Generation for Evidence Collection from Web Search Engines}

\author{Nestor Prieto-Chavana \\
  University of Sussex \\
  \texttt{n.prieto-chavana} \\
  \texttt{@sussex.ac.uk} \\\And
 Julie Weeds \\
  University of Sussex \\
  \texttt{juliewe} 
  \texttt{@sussex.ac.uk} \\ \And
 David Weir \\
  University of Sussex \\
  \texttt{d.j.weir} 
  \texttt{@sussex.ac.uk} \\}

\begin{document}
\maketitle
\begin{abstract}
It is widely accepted that so-called facts can be checked by searching for information on the Internet. This process requires a fact-checker to formulate a search query based on the fact and to present it to a search engine.  Then, relevant and believable passages need to be identified in the search results before a decision is made.  This process is carried out by sub-editors at many news and media organisations on a daily basis.  Here, we ask the question as to whether it is possible to automate the first step, that of query generation. Can we automatically formulate search queries based on factual statements which are similar to those formulated by human experts? Here, we consider similarity both in terms of textual similarity and with respect to relevant documents being returned by a search engine.  First, we introduce a moderate-sized evidence collection dataset which includes 390 factual statements together with associated human-generated search queries and search results.  Then, we investigate generating queries using a number of rule-based and automatic text generation methods based on pre-trained large language models (LLMs).  We show that these methods have different merits and propose a hybrid approach which has  superior performance in practice.
\end{abstract}



\input{ev_coll/intro}

\input{ev_coll/related}

\input{ev_coll/dataset}

\input{ev_coll/methods}

\input{ev_coll/experiments}

\input{ev_coll/results}

\input{ev_coll/summary}

\section*{Limitations}
Due to practical constraints, our dataset only contains one target URL annotation for each claim. However, it stands to reason that there may be more equally useful evidence pages available on the web. Thus, our results could be considered a lower bound on how the methods we present actually perform: the results for failed samples may as well be returning valid evidence, it's simply not the evidence selected by our annotators. 
Another limitation is the time-sensitive nature of the evidence. For some of the claims included in the dataset, the evidence was only valid for a limited amount of time. The relevance of selected target URLs may drop over time, and the information in those claims may also change over time (i.e., a relevant person's job may change, a statistic can vary). This could make it difficult to reproduce the results shown here after some time has passed.


\section*{Acknowledgements}
We would like to thank The Times and Sunday Times, for sharing their data for the creation of this dataset, as well as their sub-editing team for sharing their insights into the fact-checking process. We would also like to thank the anonymous reviewers for their feedback and comments.

\bibliography{anthology, custom}
\bibliographystyle{acl_natbib}

\appendix
\input{ev_coll/appendix}

\end{document}

%% file: ev_coll/intro.tex
\section{Introduction}
\label{sec:intro}
The need for fact-checking, i.e., to check the veracity of statements that are being presented as facts, has become an integral part of everyday life in our digital age.  Fact-checking may be carried out by the presenters of the facts, e.g., as part of an internal sub-editing process at a news organisation, or externally by the receivers of those facts.  In either case, the collection of evidence is a key step in the fact-checking process. Factual information is required to make informed decisions on the veracity of claims.  Evidence can come from a variety of sources, many of which make their information freely available on the Internet.

Search engines have become the default way for people to interact with the internet. Whilst shadowing sub-editors at a major UK newspaper performing their fact-checking duties, we found that a web search was often the first step for locating verifying evidence.  Then, sub-editors would either select pages from the search results to read or, if no relevant pages were found, they would adjust their query, making additions or other modifications, and search again until a suitable evidence page was found. 

We set out to replicate this behaviour by developing an evidence collection module for an automated system. This module takes care of generating search queries based on sentences where claims have been identified, and collecting the text from these pages. Selecting the relevant snippets and assessing the veracity of claims is beyond the scope of this paper. However, coming up with the right search queries is a non-trivial task. Long sentences may have more than one potential fact being stated. It can be difficult to identify the actual fact being verified by the user. Thus the query generation technique used is a key component of the evidence collection module. 

In this work, we analyse a number of techniques for conditional text generation and compare their performance to a number of rule-based generation baselines. We evaluate the automatically generated queries with respect to their similarity to human-crafted search queries, both in terms of textual similarity and in terms of the search results produced when executed on a search engine. 

Our contributions are threefold.  First, we introduce an evidence collection dataset specific to the fact-checking application for supporting sub-editing.  Second, we demonstrate how conditional text generation techniques can be used effectively to create useful search engines for the collection of evidence both alone and in combination with rule-based approaches.  Third, we demonstrate that similarity to human-created search queries is a useful proxy for determining whether an automatically generated query will be effective in retrieving the same evidence.  However, there are non-trivial cases when seemingly different search queries can result in the same evidence being collected. 

This paper is organised as follows: Section~\ref{sec:related} discusses related work in evidence retrieval for fact-checking and prompting/in-context learning. Section~\ref{sec:dataset} introduces the dataset we annotated for the query generation task. Section~\ref{sec:methods} explains the methods we use for conditional generation and our evaluation approach. Section~\ref{sec:experiments} describes the experiments. Section~\ref{sec:results} presents our results, and section~\ref{sec:conclusion} closes with conclusions and future work.

%% file: ev_coll/related.tex
\section{Background and Related Work}
\label{sec:related}
In this section, we review the literature on evidence collection for fact-checking (Section \ref{sec:related_evidence}) and prompting large language models / in-context learning (Section \ref{sec:related_prompting}).

\subsection{Evidence Collection in Fact-Checking Works}
\label{sec:related_evidence}

Relatively little research has been done in fact-checking using the entirety of the web as the source, as is done in this work. In \citet{popat-etal-2016-credibility}, the authors perform a credibility assessment of claims collected from the Snopes\footnote{https://www.snopes.com} fact-checking website, and from Wikipedia, classifying claims as \textit{true} or \textit{fake}. To collect evidence articles from the web, the authors use the full claim text on the Google search engine and scrape the first three pages of results from each. The authors use a reliability score for evidence pages based on the website's AlexaRank and PageRank~\citep{page1999pagerank}, as well as a series of lexicons to create a representation of the linguistic style of these articles. They found that the source reliability score had little effect on the outcomes.

\citet{karadzhov_fully_2017} presents an approach using web search as a source of evidence for fact-checking, using both Bing and Google as search engines. Search queries are constructed using a combination of rules: adding the top-ranked verbs, nouns and adjectives using TF-IDF, as well as adding all named entities present in the claim. Queries are limited to 5-10 tokens. They evaluate their approach in the rumour detection dataset published by \citet{ma-etal-2016-detecting}. They report finding no practical difference in performance from using Bing or Google. They also found almost no difference when using the full text of the page as evidence versus only the snippet delivered by the search engine.

The FEVER task \citep{thorne-etal-2018-fever} consists of the classification of claims into Supported, Refuted and Not Enough Info, based on evidence contained in Wikipedia. Systems are evaluated not only on their ability to produce the correct classification but also on the retrieval of the correct evidence. Participants in FEVER have used a variety of approaches to evidence retrieval, sometimes including the use of web search or similar search APIs. For example, \citet{hanselowski-etal-2018-ukp} extract Noun Phrases from the claim text and use them as search queries on the Wikipedia API, which finds matching articles using the Wikipedia search engine. This approach resulted in the highest evidence recall of all participants in the first edition of the shared task and was re-used by participants in later editions. In FAKTA~\citep{nadeem-etal-2019-fakta}, in addition to using the 2017 Wikipedia dump for evidence, the authors use the Google Custom Search API to search specific websites for evidence, classified by their reliability by the Media Bias/Fact Check website\footnote{https://mediabiasfactcheck.com}. Queries are created by only considering the verbs, nouns, adjectives, and named entities present in a claim.

In \citet{fan-etal-2020-generating}, the authors introduce a dataset geared towards the generation of fact-checking briefs, intended to present information that can assist fact-checkers. This includes QA briefs, which extract specific questions and answers based on the claim. To generate these, the authors train a text-to-text model, BART~\citep{lewis-etal-2020-bart}, to automatically generate questions based on the claim of the text. They then train a QA model to generate answers to these questions, based on the question text and the source document used to originally fact-check the source claims. Our work differs from this in that the claim text is used to generate search queries that are then applied directly to a search engine, in order to find evidence documents from the web. 

\subsection{Prompting and In-context Learning}
\label{sec:related_prompting}

  A recent paradigm shift in the usage of large language models (LLMs) has been referred to as the “pre-train, prompt, and predict” approach \citep{liu_pre-train_2021}.  The intuition behind it is to modify the task at hand to approximate more closely the model pre-training task. Since LLMs are pre-trained with language modelling tasks, these modifications have consisted of changing the input, so it more closely resembles natural language. Depending on the task at hand this has been done through the use of templates, for example, to turn classification tasks into cloze tasks \citep{petroni-etal-2019-language}. It can also be done through the use of prompts \citep{brown_language_2020} which are short textual explanations of the task added to the model input. The line of research focused on identifying effective prompts is known as prompt engineering \citep{liu_pre-train_2021}. In fact, the main motivation for the T5 model \citep{raffel_exploring_2020} was the reformulation of traditional NLP tasks into text-to-text (text as input, and text as output) tasks, so that other tasks can be attempted using a prompting approach.

An extension to this approach is to also include solved examples of the task the model needs to perform. This is known as in-context learning \citep{brown_language_2020}, also referred to as few-shot learning. Using in-context examples has been shown to produce results that are competitive with the fine-tuning approach in tasks such as  translation, question-answering, and cloze tasks. Although not able to surpass fine-tuning in performance, in-context learning has the advantage of requiring smaller amounts of data and computing resources, as the model is only performing inference.

The selection of examples to add to the context is one of the key elements of this method, as it can have a large effect on the model's performance\citep{lu-etal-2022-fantastically}. These examples should reflect the task that the model is expected to perform. Examples can be randomly sampled from a training dataset \citep{brown_language_2020}, or a smaller held-out development dataset. However, in a truly limited data setting, this may not be a viable approach. An alternative could be to sample examples from closely-related tasks. However, \citet{perez_true_2021} have argued against such use of out-of-domain data.

%% file: ev_coll/dataset.tex
\section{Dataset}
\label{sec:dataset}

In this section, we describe the dataset that we have compiled and annotated in this research. The dataset is composed of 390 claims containing verifiable information. Each claim is accompanied by a target URL, an evidence page containing the information required to verify the claim, along with a target search query, which was used to find the target URL. The source of claims is an archive of news articles published in The Times and Sunday Times, and includes content from 20 different sections. A random
sample of 100 articles published between May and June of 2021 were selected from this
collection for annotation.

During the annotation process, articles were first split into sentences using the Spacy\footnote{https://spacy.io/} library. Second, annotators read through the articles sentence by sentence, assessing whether each sentence contained verifiable information. Only sentences labelled as verifiable progressed to the next annotation steps. Third, annotators looked for relevant evidence that would allow the verification of claims. In order to do this, they used the Bing search engine to find relevant web pages. The motivation for this process was to follow the usual approach of a fact-checker looking for information based on a claim, and not place any artificial constraints on the format of the query used. The only limitation set was that the chosen web page should be found on the first page of search results. Annotators could construct the search queries using information contained in the claim sentence as well as additional context from other sentences of the article if necessary. If no relevant page was found with their initial query, they could modify them and retry as many times as needed. Annotators would review results and select the highest-ranking page that contained the necessary information to verify the claim.  Once an evidence web page was found, the claim sentence would be annotated with the final search query used to find it. Finally, they would be annotated with the URL of the selected evidence web page. 
Working with historic article data presented some challenges, including the fact that many of the claims are echoed through other media outlets, often saturating the search results provided by search engines. Throughout the annotation process, annotators were instructed to prioritise first-hand sources, and avoid using other news/media sources where possible. If news/media sources were the only apparent verification, priority was given to those sources which had earlier publication dates than the checked article.  

During our experiments, we found that search results from the Bing search engine were variable, even when using the same query. The original dataset contained more than the current 390 claims. Repeated executions of the target search queries produced varying results, and the target URLs were not always present in the search results. This was a concern since this would be one of the metrics used to evaluate the quality of the generated queries. In order to ensure that the dataset would only include samples where the target URL was able to be found using the corresponding target search query, we executed all target search queries on Bing a total of 12 times on 3 different days and collected the results. Only the samples where the target URL was found in the search results in 50\% or more of the 12 executions were kept in the dataset. This resulted in 390 of the samples remaining in the dataset. All fine-tuning and evaluation are done on these 390 samples. Table \ref{tab:ec-dataset-examples} shows some examples of claims, each with their corresponding target search query and target URL.

\begin{table*}
\centering
\begin{tabular}{p{0.35\textwidth} p{0.2\textwidth} p{0.35\textwidth}}
\cline{1-3}
Claim & Target Search Query & Target URL \\
\cline{1-3}
Liz Truss, the international trade secretary, said that membership of the CPTPP represented "a huge opportunity for Britain". & Liz Truss international trade secretary & https://en.wikipedia.org/wiki/\newline liz\_truss \\

The OECD raised its global growth forecast from 5.6 per cent to 5.75 per cent for this year and from 4 per cent to nearly 4.5 per cent for 2022. & oecd global gdp growth forecast may 2021 & https://www.oecd.org/economic-outlook/may-2021 \\

Galileo, the European system, operates at a height of 23,000km, but Britain is being evicted from Galileo because of Brexit. & galileo operating height & https://en.wikipedia.org/wiki/\newline galileo\_(satellite\_navigation) \\

\end{tabular}
\caption{Sample of claims, target search queries and corresponding target URL in the dataset.}
\label{tab:ec-dataset-examples}
\end{table*}

%% file: ev_coll/methods.tex
\section{Methodology}
\label{sec:methods}
In this section, we first define the query generation task \ref{sec:methods_query}.  We then outline rule-based \ref{sec:methods_rules} and conditional text generation \ref{sec:methods_cond} approaches to solving it and the evaluation methods \ref{sec:methods_eval}  we have employed.      
\subsection{The Query Generation task}
\label{sec:methods_query}
The query generation task is defined as follows: each sample is composed of a text claim \(c\), a human-created query \(q\)  and a target url \(url\), which contains the evidence required to validate \(c\). Given the text of the claim \(c\), generate a search query \(q'\) that, when executed on a search engine, will produce \(url \) within the first page of results.  We evaluate the quality of \(q'\) by its similarity to \(q\) , as well as by executing it on the Bing search engine and looking for \(url\) in the results. 

\subsection{Rule-based Query Generation}
\label{sec:methods_rules}
Rule-based query generation approaches take into account the syntactic features of the tokens in claim \(c\) to decide whether they should be included in query \(q\).  We consider 3 approaches described below:

\noindent\textbf{Verbatim}: This method uses the full text of the claim as query. The search engine will likely find close paraphrases of the text in \(c\). 

\noindent\textbf{Named Entities}: All named entities identified in the sentence are concatenated together. This approach is expected to find results where the entities are mentioned together, and which may be stating similar relations between them, but may not necessarily agree with the claims being made in the source sentence.

\noindent\textbf{Noun Phrases}: All noun phrases identified in the sentence are concatenated together. This is similar to the above; however, as the definition of a noun phrase is broader than that of a Named Entity, it incorporates additional content from the claim.

\subsection{Conditional Text Generation}
\label{sec:methods_cond}
We model the query generation task as conditional text generation. That is, a text generation model is input with the full text of the claim \(c\) together with any additional contextual information, and outputs the query \(q'\). We experiment with three conditional text generation approaches:

\noindent\textbf{Zero-shot learning}: 
A prompt template is used to add a prefix to the claim \(c\) before it is input to the text generation model.   We experiment with different prompts, described in Section \ref{sec:experiments_prompts}.

\noindent\textbf{Few-shot learning}: In few-shot learning, the model is shown a small number of examples before attempting the task.  It is done by pre-pending these examples directly before the main task input. 

\noindent\textbf{Conventional Fine-tuning}: Finally, we also experiment with conventional fine-tuning. In this setting, the T5 model is fine-tuned receiving the claim sentence \(c\) as input and human-created query \(q\) as target. Due to the limited size of the dataset, we use 4-fold cross-validation. For each fold, the model is fine-tuned on ~75\% of the data and tested on the remaining 25\%. We average the evaluation metrics over the 4 test sets.

\subsection{Evaluation}
\label{sec:methods_eval}
We evaluate the different query generation methods in two ways: (1) on the generated queries' similarity to the target queries; and (2) on the generated queries' ability to return the target URLs. 

\subsubsection*{Query Similarity Evaluation}

Our reasoning for using this evaluation is that a model that produces the same query as the human annotator is a high-quality model. It further stands to reason that an identical query has a higher probability of producing the same search results than a different one. This is true even when considering the search result inconsistencies described before.

\noindent\textbf{Rouge}: We use Rouge \citep{lin-2004-rouge} to evaluate the automatically generated queries' textual similarity to the target queries. This metric is widely used to evaluate model performance in tasks such as summarisation and translation. We use Google Research's python implementation of Rouge\footnote{https://github.com/google-research/google-research/tree/master/rouge}. Since both the target and the automatically generated queries are quite short, we instead use character-based Rouge, ie. we replace the standard tokeniser with simply splitting sentences into characters. We report the Rouge-1, Rouge-2 and Rouge-L metrics for all experiments.

We also considered using the Levenshtein or edit distance \citep{levenshtein1966binary} between queries.  Whilst, theoretically, character-based Rouge scores should be less affected by the re-ordering of words in queries, we found in practice that Levenshtein distance scores were highly correlated with the Rouge-L scores, so we do not include them here. However, Levenshtein distance is used to select the optimal samples to be used as in-context examples in few-shot learning (see Section \ref{sec:experiments_fewshot}).

\subsubsection*{Search Result Evaluation}

We execute the queries via the Bing search API\footnote{https://www.microsoft.com/en-us/bing/apis/bing-web-search-api} and collect the top 10 results, corresponding to the first page in a browser. We evaluate whether the target URL appears in the search results. 

There are a number of complications to this evaluation. The first is the consistency of the search results: during our experiments, we found that using the same query does not guarantee receiving the same results from the search engine. In order to account for this variation, we execute each query 3 times and collect the three different ranked lists of search results.
Second, this evaluation comes with a financial cost from using the search provider's APIs. This cost can quickly add up when evaluating different methods and hyperparameter configurations, both for fine-tuning and generation. This is compounded by the repeated executions required in order to improve the consistency of the results. To account for this, we use query similarity evaluation when tuning model hyperparameters, and only use search result evaluation on the final models.  We use the following metrics for search result evaluation:

\noindent\textbf{Found All Percentage (FA\%)}: The proportion of test samples where the target URL was present in the search results for all three executions.

\noindent\textbf{Found Majority Percentage (FM\%)}: The proportion of test samples where the target URL was present in the search results for the majority of executions, i.e. at least two.

\noindent\textbf{Found Once Percentage (FO\%)}: The proportion of samples where the target URL was found in the results for at least one of the executions.

\noindent\textbf{MRR}: Mean Reciprocal Rank (MRR@K) is a metric frequently used for evaluation in information retrieval tasks \citep{voorhees-tice-2000-trec}. This metric takes into consideration the highest-ranked relevant answer in the retrieved documents.  We set \(K=10\), corresponding to the first page of results of a search engine.  Each list of search results corresponding to an executed search query is considered separately and contributes to the overall MRR score.

%% file: ev_coll/experiments.tex
\section{Experiments}
\label{sec:experiments}
In this section, we describe the specific implementations of each approach used in our experiments.

\subsection{Rule-based Query Creation Baselines}

 We use the Spacy library to extract Named Entity Mentions as well as the parse tree of the text in a claim.  Table \ref{tab:querygenexamples} in Appendix~\ref{sec:appendix} shows examples of the application of the different rule-based methods.   

\subsection{Conditional Text Generation}
\label{sec:experiments_prompts}

We use the T5 language model \citep{raffel_exploring_2020}, specifically, the language generation adapted model (t5-lm-adapt) introduced in \citet{lester-etal-2021-power}. This version is trained for an additional 100K steps on the LM task, in order to overcome training limitations in the original T5 model which struggles to use uncorrupted text. We use the large checkpoint of the model which has 800 million parameters.  
We use the implementation provided by the Huggingface Transformers python library\footnote{https://github.com/huggingface/transformers}. We set the following parameters for text generation: we use beam search with 10 beams, restrict the model from repeating bigrams, and set generation early stopping to true. Since the generated queries should not be very long, we set the maximum number of new tokens to 16. 

During our experiments, we noticed that the T5 model has a tendency to reproduce parts of the prompt in the generated text. We observed this specifically in the few-shot and zero-shot learning experiments. To account for this, we add a post-processing step to the generation: if the prompt text (either from the prefix or suffix) is included verbatim in the model output, it is removed. Partial inclusions are not removed from the output.

As found in \citet{petroni-etal-2019-language,brown_language_2020, schick-schutze-2021-just}, the choice of prompt to use has a significant effect on model performance.  Out of the tasks included in the original T5 paper, we consider summarisation to be the closest to the query extraction task, and thus experiment with the ``summarize'' prompt here.  We also experiment with prompts that describe the task the models needs to perform in different ways: \textit{Generate Search Query}, \textit{Fact-check the following sentence}, \textit{Verify the following sentence} and \textit{Summarize the following sentence} prompt. We also consider shorter prompts, which only use one or two tokens to indicate the action required from the model: \textit{Search query}, \textit{Fact-check}, \textit{Verify} and \textit{Summarize}. We experiment with using the shorter prompts both as prefix and suffix to the model input.  The prompts used in our experiments are detailed in Appendix\ref{sec:appendix}.

\subsubsection{Zero-shot learning}
With this method, the prompt selection is the only thing guiding the output produced by the model. 

\subsection{Few-shot learning}
\label{sec:experiments_fewshot}
We evaluate the few-shot learning method using 4-fold cross-validation. The split is done based on the source article of the claims, meaning claims originating on the same article only appear in one of the sets, training or testing.  See Table~\ref{tab:ec-dataset-fold-sizes} in Appendix \ref{sec:appendix} for the number of records in the train and test set for each of the folds. 
At each fold, the model is evaluated on 25\% of the data, and the remaining 75\% can be used as in-context examples for the text generation model. In order to select the optimal in-context examples, we first run a one-shot learning evaluation, using each individual sample in the training set as examples, and evaluating the generated queries against the rest of the training set using the Levenshtein distance metric. The best 3 performing examples then undergo the complete evaluation, in the one, two and three-shot settings.

\subsection{Conventional Fine-tuning}

We again use 4-fold cross-validation.  Models are fine-tuned on 75\% of the claims and evaluating on the remaining 25\% each time. This split is done as before based on the source article.  For each fold, ~15\% of the training data is selected as a validation set.  The validation set is used to evaluate the model performance at the end of each fine-tuning epoch. The model is evaluated based on the Levenshtein similarity ratio to the target search queries of the validation set. At the end of the fine-tuning process, the best performing checkpoint of the model is saved. Each model is fine-tuned for 10 epochs on a single GeForce RTX 3090 GPU. Following \citet{raffel_exploring_2020}, we use the Adafactor optimiser, with a learning rate of 1e-3 and max input sequence length of 512. We use the same generation parameters as for zero-shot and few-shot learning.

%% file: ev_coll/results.tex
\section{Results and Discussion}
\label{sec:results} 

\begin{table*}
\centering
\begin{tabular}{|l|c|c|c|c|c|c|c|c|}
\hline
Generation & & & & & & & \\ Method & R-1 & R-2 & R-L & FA\%  & FM\% & FO\% & MRR \\
\hline
Verbatim Claim & 0.343 & 0.294 & 0.314 & 15.1 & 17.2 & 18.2 & 0.114 \\
Noun Phrases & 0.455 & 0.368 & 0.396 & 19.7 & 20.5 & 22 & 0.145 \\
Named Entities & \textit{0.577} & \textit{0.423} & \textit{0.484} & \textit{30.0} & \textit{32.3} & \textit{33.8} & \textit{0.217} \\
\hline
zero-shot &	0.533 &	0.388 &	0.438 & 25.1 & 26.2 &	29 &	0.181 \\
one-shot &	0.546 &	0.339 &	0.418 & 19.0 & 20.5 &	23.1 &	0.15 \\
two-shot &	0.582 &	0.382 &	0.46 & 25.6 & 26.4 &	28.2 &	0.191 \\
three-shot &	0.593 &	0.402 &	0.474 & 30.3 & 31.5 &	32.8 &	0.225 \\
fine-tuned &\textbf{0.646} & \textbf{0.472} & \textbf{0.56} & \textbf{41.0} & \textbf{41.8} & \textbf{42.6} & \textbf{0.321} \\
\hline
\end{tabular}
\caption{Performance metrics for each query generation method and rule-based baselines. For each generation method, the best-performing template is shown.} 
\label{tab:ev-coll-overall-best}
\end{table*}

Table~\ref{tab:ev-coll-overall-best} shows the results for the rule-based baselines, as well as the top performing prompt for the zero-shot, few-shot and fine-tuning methods, measured by the FM\%. 
The effect of prompt selection on results is discussed in Appendix~\ref{sec:appendix}.  An analysis of the results shows that FA\%, FM\% and FO\% are not very far apart. On average, the difference between the three metrics is ~4\% across all experiments. In other words, if the target result was found at least once in the results, it's very likely to come up most of the time. However, the fact that the metrics diverge confirms that web search results are non-deterministic. They can also change in a short time span: all Bing searches for each query were done immediately one after another, yet still frequently produced different results. 

We also observed that web pages seem to drop out of results quite quickly. Recall that at the time of dataset creation, the target queries were validated to confirm they returned the target URLs. Only those queries that returned the target URL the majority of the time were included in the dataset. Later, during our experiments, we tested using the target queries for collection. As shown in Table~\ref{tab:target-queries-results-websearch}, only 93.6\% of the target queries returned the target URL in the results, while 95.1\% returned it at least once. This means in the time between dataset creation and running these experiments (approximately two months), the target URL dropped completely from the results for almost 5\% of the samples. This constitutes an upper bound on how well any system can perform.

\begin{table}
\centering
\begin{tabular}{|l|c|c|c|c|}
\hline
Generation & & & & \\
Method & FA\% & FM\% & FO\% & MRR \\
\hline
Target Queries&92.1&93.6&95.1&0.697\\
\hline
\end{tabular}
\caption{Web search evaluation results using target queries.} 
\label{tab:target-queries-results-websearch}
\end{table}

Since we evaluate the generated queries in relation to textual similarity with the target queries, as well as the quality of the search results, we assess the correlation between these two evaluation methods. The results are shown in Table~\ref{tab:metrics-correlation} and confirm that similarity to the target query is a very good indicator of the potential quality of search results. The Rouge-2 metric has a very high correlation with the FA\%, FO\% and FM\% metrics, while Rouge-L shows the highest correlation with the MRR score. 

\begin{table}
\centering
\begin{tabular}{|l|c|c|c|c|}
\hline
& FA\% & FM\% & FO\% & MRR \\
\hline
Rouge-1  & 0.842 & 0.831 & 0.817 & 0.848 \\
Rouge-2  & 0.976 & \textbf{0.98} & \textbf{0.984} & 0.972 \\
Rouge-L  & \textbf{0.977} & 0.972 & 0.966 & \textbf{0.979} \\
\hline
\end{tabular}
\caption{Correlation between similarity-based and search-based metrics.} 
\label{tab:metrics-correlation}
\end{table}

However, a low similarity score does not necessarily mean that the query will not return the target URL. While analysing the results, we found many cases where queries with low Rouge scores still produced the desired search results. In the majority of these cases, the target URL was a Wikipedia page, which meant that the inclusion of a single named entity in the search query was likely enough for the search engine to produce the related page. Conversely, we also found generated queries with very high similarity to the target queries (based on the Rouge-2 metric), that nevertheless failed to return the target URL.

Returning to the results in  Table~\ref{tab:ev-coll-overall-best}, we note that the simple named entities baseline approximates the target queries most closely of our rule-based baselines. This suggests that a high proportion of human-created queries include named entities. 

Of the in-context generation methods, three-shot learning achieves competitive performance with the NE baseline, but it fails to surpass it. None of the other in-context methods come close to the baseline performance. Meanwhile, fine-tuning outperforms the named entity baseline. This suggests that while named entities are an important element of the target queries, there are other factors that go into making an effective query. For example, these could be selecting the correct named entity when several are present in the text, and including other words which are not named entities.

To investigate the differences in performance between the various generation methods, we consider the specific samples that each of the top-two methods are able to solve correctly. Table~\ref{tab:ner-ft-twoway} shows the overlap between the correct samples for the NE baseline and the fine-tuning method. While the NE baseline is outperformed by the fine-tuning method, there is a significant proportion of samples (30 out of a possible 227) that NE gets correct and fine-tuning does not. This points to the fact that different methods have different strengths, and the optimal way to employ them would be to combine those strengths. 

\begin{table}
\centering
\begin{tabular}{|l|c|c|c|}
\hline
& NE Matches & NE Fails & Total \\
\hline
Fine-tuning	& & & \\
Matches & 96  &  67  & 163 \\
\hline
Fine-tuning	& & & \\
Fails	& 30  & 197  & 227  \\
\hline
Total &	126	 & 264  & 390 \\
\hline
\end{tabular}
\caption{Overlap of samples for which the NE baseline and the fine-tuning generation method produce the target URL in the search results.} 
\label{tab:ner-ft-twoway}
\end{table}

\begin{table*}
\centering
\begin{tabular}{|l|c|c|c|c|}
\hline
Generation Method & FA\% &  FM\% & FO\% & MRR \\
\hline
NE + Fine-tuning & 45.1 & 46.9 & 47.9 & 0.332 \\
NE + Three-shot + Fine-tuning & 44.9 & 46.7 & 47.9 & 0.345 \\ 
All Rule-based (NE + NP + Verbatim) & 35.4 & 37.9 & 41.3 & 0.227 \\ 
All Rule-based + Fine-tuning & 47.9 & 49.2 & 50.5 & 0.331 \\ 
All Rule-based + Three-shot + Fine-tuning & 47.4 & 49.7 & 51.3 & 0.356 \\ 
\hline
\end{tabular}
\caption{Web search evaluation results combining different methods using the Borda count.} 
\label{tab:combined-results-websearch-borda}
\end{table*}

In order to explore the potential for an ensembled system further, we use a modified Borda count method to combine the rankings of each of the query generation methods, while prioritising the most individually effective. We produce a combined top 10 ranking of search results, and score this using the same metrics as for individual methods. We use this approach to combine the top-two performing query generation methods (NE and Fine-tuning) and the top-three methods (NE, Fine-tuning and Three-shot). We also experiment with combining all rule-based baseline methods. As can be seen in Table~\ref{tab:combined-results-websearch-borda}, an ensemble of different methods can outperform any of the single methods being combined.

\subsection{Error Analysis}

Even when combining the top-performing generation methods, the performance metrics still show much room for improvement.
In order to identify the best opportunities for improvement, we analysed a sample of the fine-tuning model output that failed to produce the target URL, and classify them. A detailed breakdown is shown in Table~\ref{tab:ft-error-analysis} in Appendix~\ref{sec:appendix}. The most common error mode is for the model to fail to include key terms in the query, which change the nature of the information being sought. The second most common corresponds to cases where the target query included terms or context external to the claim sentence. This could potentially be solved by including text from the article title, or previous sentences, in the model input. The third most common cause of error is the model focusing on the wrong named entity when multiple are present in the claim text. With the same frequency, we also found cases of hallucination, where the output of the model did not take into account the text input, and instead, it produced text from what appeared to be training samples. The next failure mode is where the model simply reproduced the claim text verbatim up to its token limit. Finally, we found a small percentage of cases where, while the query was different from the target, it appeared to cover the key terms necessary to get the correct results. In these cases, it may be due to a change in the website structure, or simply due to the inherent variation in web search engine results.

%% file: ev_coll/summary.tex
\section{Conclusion}
\label{sec:conclusion}
In this paper, we have introduced an evidence collection dataset specific to fact-checking in the context of sub-editing news articles. We have shown that conditional text generation approaches can improve on rule-based baselines on their ability to produce the correct search results. We have also explored the effect of ensembling several of these approaches, and shown that this outperforms any single approach. 
We have also analysed the shortcomings of different generation methods, and found areas of potential improvement. In the future, there is scope to experiment with methods of including additional context in the automated model input. There is also scope to experiment with more involved ways of combining different methods' input, e.g., the output of rule-based baselines could be used to enhance the inputs for the automated generation methods. Finally, further data annotation or data augmentation methods might be used to overcome the challenges of having a dataset which is relatively limited.

%% file: ev_coll/appendix.tex
\section{Appendix}
\label{sec:appendix}

Table~\ref{tab:ev-coll-prompt-templates} Shows the different templates used for prompting in the zero-shot and few-shot generation methods. The same templates were applied during fine-tuning.
The scores reported in Table~\ref{tab:combined-results-websearch-borda} were achieved when using template-03 for zero-shot learning, template-06 for one-shot learning, template-05 for two and three-shot learning, and no-prompt for fine-tuning. In general, prompts with longer task descriptions were more useful for zero-shot learning. Meanwhile, short suffix prompts were produced the best results for few-shot learning. When using shorter prompts, having them as a suffix to the input produced better results, while short prefix prompts produced poor results, especially in the few-shot setting. For fine-tuning, the choice of prompt had very little effect, while the best performance was achieved when using no prompts at all.

\begin{table*}
\centering
\begin{tabular}{|l|l|p{2cm}|p{2cm}|p{4cm}|}
\hline
Format & Template ID&Prefix&Suffix&Zero-shot example\\
\hline
no-prompt & no-prompt & - & - & Gerry Ford, 63, founded the business in 1997. \\
\hline

Long-explanation & template-01 & Generate search query: & Search query: &  \textit{Generate search query:} Gerry Ford, 63, founded the business in 1997. \textit{Search query:} \\
 &template-02 & Fact-check the following sentence: & Fact-check: &  \textit{Fact-check the following sentence:} Gerry Ford, 63, founded the business in 1997. \textit{Fact-check:}  \\
 &template-03 & Verify the following sentence: & Verify: &  \textit{Verify the following sentence:} Gerry Ford, 63, founded the business in 1997. \textit{Verify:}  \\
 &template-04 & Summarize the following sentence: & Summarize: &  \textit{Summarize the following sentence:} Gerry Ford, 63, founded the business in 1997. \textit{Summarize:}  \\
\hline
Short-suffix & template-05 & - & Search query: &  Gerry Ford, 63, founded the business in 1997. \textit{Search query:}  \\
& template-06 & - & Fact-Check:  &  Gerry Ford, 63, founded the business in 1997. \textit{Fact-Check:}  \\
 & template-07 & - & Verify: &  Gerry Ford, 63, founded the business in 1997. \textit{Verify:}  \\
 & template-08 & - & Summarize: &  Gerry Ford, 63, founded the business in 1997. \textit{Summarize:}  \\
\hline
Short-prefix & template-09 & Search query: & - &  \textit{Search query:} Gerry Ford, 63, founded the business in 1997. \\
& template-10 & Fact-Check: & - &  \textit{Fact-Check:} Gerry Ford, 63, founded the business in 1997. \\
& template-11 & Verify: & - &  \textit{Verify:} Gerry Ford, 63, founded the business in 1997. \\
& template-12 & Summarize: & - &  \textit{Summarize:} Gerry Ford, 63, founded the business in 1997. \\

\hline
\end{tabular}
\caption{Prompt templates used for zero-shot, few-shot and fine-tuning methods.}
\label{tab:ev-coll-prompt-templates}
\end{table*}

Each method's sensitivity to prompt choice was highly varied. The few-shot setting showed to be the most sensitive, with experiments using no prompts or short-prefix type prompts resulting in very low performance. This effect was not as noticeable for zero-shot learning, while in fine-tuning, the prompt choice had a negligible effect. Table~\ref{tab:prompt-choice-var} shows the average performance for each method across all prompt templates, as well as the standard error. We also show the same metrics for few-shot while removing the results using no prompts or short-prefix style prompts. 

\begin{table*}
\centering
\begin{tabular}{|l|c|c|}
\hline
Method & R-2 & FM\% \\
\hline
zero-shot & 0.340 (0.008) & 19.7 (1.1) \\
fine-tuning & 0.454 (0.004)	& 38.4 (0.5) \\
\hline
one-shot & 0.263 (0.014)	& 12.1 (1.7) \\
two-shot & 0.285 (0.026)	& 16.2 (2.9) \\
three-shot & 0.297 (0.027)	& 18.8 (3.2) \\
\hline
one-shot* & 0.299 (0.009)	& 16.5 (1.1) \\
two-shot* & 0.359 (0.004)	& 24.4 (0.8) \\
three-shot* & 0.372 (0.005)	& 27.9 (0.7) \\
\hline
\end{tabular}
\caption{Variability of results by prompt choice. Numbers are the average of metrics across experiments using different prompt templates. Numbers in parenthesis are standard error. Methods marked with an asterisk exclude experiments using no prompt or short-prefix prompts.}
\label{tab:prompt-choice-var}
\end{table*}

Table~\ref{tab:querygenexamples} shows the output of applying the rule-based baselines to the example sentence: \textit{Netanyahu didn't become Israel's longest-serving prime minister by mistake.} The fact under verification is whether Netanyahu was indeed the longest-serving prime minister of Israel.

\begin{table*}
\centering
\begin{tabular}{|p{4cm}|p{8cm}|}
\hline
Generation Method&Query\\
\hline
Verbatim&Netanyahu didn't become Israel's longest-serving prime minister by mistake.\\
\hline
Named Entities &Netanyahu, Israel\\
\hline
Noun Phrases& Netanyahu,Israel's longest-serving prime minister, mistake \\ 

\hline
\end{tabular}
\caption{Output of different rule-based query generation baselines.} 
\label{tab:querygenexamples}
\end{table*}

Table~\ref{tab:ec-dataset-fold-sizes} shows the size of training, dev and test sets derived using 4-fold cross-validation. The split was done based on the source article of the claims.

\begin{table*}
\centering
\begin{tabular}{|l|c|c|c|}
\hline
Fold & Train & Dev & Test \\
\hline
K0 & 245 & 45	& 100 \\
K1 & 246	& 48	& 96 \\
K2 & 243	& 48	& 99 \\
K3 & 248	& 47	& 95 \\
\hline
\end{tabular}
\caption{Number of samples in fine-tuning data splits.}
\label{tab:ec-dataset-fold-sizes}
\end{table*}

Table~\ref{tab:ft-error-analysis} contains the results and examples from the error analysis of a sample of fine-tuned model predictions. 

\begin{table*}
\centering
\begin{tabular}{|p{4cm}|p{6cm}|c|}
\hline
Reason  & Example & Percentage \\
\hline
Query missing key information/term & Claim Text: Meanwhile, Hostelling Scotland — the YHA’s counterpart north of the border, turning 90 this year — began a phased reopening of its 60-plus hostels two days ago.  	& 30\% \\
Query required context outside claim& Claim Text: The opening ceremony, in the National Stadium in Tokyo, is set for July 23.	&	26.7\% \\
& Target Query: tokyo olympics opening ceremony & \\
& Generated Query: National Stadium in Tokyo & \\
& Target Query: hosteling scotland history & \\
& Generated Query: hostelling scotland & \\
Extracted wrong named entity & Claim Text: Priti Patel is considering withdrawing the UK from part of the European Social Charter that gives citizens of 26 countries a £55 discount on application fees for most worker visas.	&	16.7\% \\
& Target Query: european social charter fee exemption & \\
& Generated Query: Priti Patel & \\
Hallucination/Memorised from training data&	Claim Text: For the ride-hailing giant, that judgment in February could be the beginning of a wider reckoning that lands it with a £2 billion-plus VAT bill.&	16.7\% \\
& Target Query: uber february £2billion & \\
& Generated Query: mup scotland & \\
Recreated claim text& Claim Text: Colorado and Oregon have followed suit with similar lotteries and on Thursday California announced an even larger draw, with ten prizes worth \$1.5 million.	&	6.7\% \\
& Target Query: california vaccine lottery amount & \\
& Generated Query: Colorado and Oregon have followed suit with similar lotteries & \\
Query looks good & Claim Text: Chris James, founder of the impact investing firm Engine No. 1, scored a stunning victory over the biggest beast in the Big Oil jungle last week: Exxon Mobil.	&	3.3\% \\
& Target Query: chris james engine no 1 & \\
& Generated Query: Chris James founder of the impact investing firm Engine No. & \\
\hline
\end{tabular}
\caption{Fine-tuned model output error analysis.} 
\label{tab:ft-error-analysis}
\end{table*}